\documentclass[sigconf]{acmart}

\setcopyright{none}
\renewcommand\footnotetextcopyrightpermission[1]{}
\setcopyright{none}
\makeatletter
\renewcommand\@formatdoi[1]{\ignorespaces}
\makeatother

\usepackage{booktabs} 
\usepackage{xargs}
\usepackage{listings}
\usepackage{enumitem}
\usepackage{array}

\newenvironment{conditions}[1][where:]
  {#1 \begin{tabular}[t]{>{$}l<{$} @{${}={}$} l}}
  {\end{tabular}\\[\belowdisplayskip]}

\begin{document}


\title{Efficient Clustering from Distributions over Topics}

\author{Carlos Badenes-Olmedo}
\orcid{0000-0002-2753-9917}
\email{cbadenes@fi.upm.es}
\affiliation{%
  \department{Ontology Engineering Group}
  \institution{Universidad Polit\'ecnica de Madrid}
  \city{Boadilla del Monte}
  \country{Spain}
  }
\author{Jos\'e Luis Redondo-Garc\'ia}
\orcid{0000-0002-7413-447X}
\email{jluisred@amazon.com}
\affiliation{%
  \institution{Amazon Research}
  \city{Cambridge}
  \country{UK}
}
\author{Oscar Corcho}
\orcid{0000-0002-9260-0753}
\email{ocorcho@fi.upm.es}
\affiliation{%
  \department{Ontology Engineering Group}
  \institution{Universidad Polit\'ecnica de Madrid}
  \city{Boadilla del Monte}
  \country{Spain}}

\renewcommand{\shortauthors}{C. Badenes-Olmedo et al.}

\begin{abstract}

There are many scenarios where we may want to find pairs of textually similar documents in a large corpus (e.g. a researcher doing literature review, or an R\&D project manager analyzing project proposals). To programmatically discover those connections can help experts to achieve those goals, but brute-force pairwise comparisons are not computationally adequate when the size of the document corpus is too large. Some algorithms in the literature divide the search space into regions containing potentially similar documents, which are later processed separately from the rest in order to reduce the number of pairs compared. However, this kind of unsupervised methods still incur in high temporal costs. In this paper, we present an approach that relies on the results of a topic modeling algorithm over the documents in a collection, as a means to identify smaller subsets of documents where the similarity function can then be computed. This approach has proved to obtain promising results when identifying similar documents in the domain of scientific publications. We have compared our approach against state of the art clustering techniques and with different configurations for the topic modeling algorithm. Results suggest that our approach outperforms ($>0.5$) the other analyzed techniques in terms of efficiency.

\end{abstract}

%
%


%
%


\thanks{This work is supported by project Datos 4.0 with reference TIN2016-78011-C4-4-R, financed by the Spanish Ministry MINECO and co-financed by FEDER.}

\maketitle

\section{Introduction}
\label{sec:introduction}

Given the huge amount of information about any domain that is being produced or captured daily, it becomes crucial to provide mechanisms for automatically identifying the elements that can bring value for the involved agents (general consumers, experts, companies, investors...) and discard the noisy, non-relevant information. Much of the information is presented in the form of textual documents, making necessary for experts to browse through many of these texts to find relevant data. A way to explore the knowledge inside collection of documents is by moving from one information element to another based on certain criteria that relates them. This approach requires to calculate a similarity matrix with all possible comparisons between elements, so we can later select the most pertinent ones. Since computing a $n \times n$ matrix takes $O(n^2)$ time, obtaining all possible pairs of similarities in a large collection of documents can be unfeasible because of the exponential cost of comparing every pair of elements.

Our work is derived from a real need in the domain of digital libraries, where we targeted the task of finding relations among texts based on similar content inside a corpus containing 7,487 digital books and 97,532 chapters (104,960 documents in total). Since the time consumed in calculating the similarity score between two documents was $t=7.62*10^{-4}$ seconds in a 15x CPU@2.30Ghz and 64GB RAM server, the total time to compute all combinations over the whole corpus went up to around 5 days. Considering that other tasks leveraging the entire collection such as training a Topic Model only required 48 minutes to be executed, calculating the similarity scores between pairs of documents becomes a significant bottleneck when making sense of big collections of documents.

One possible way of finding similarity-based links between pair of documents, is to 1) process the items following different annotation techniques (entities, keywords, etc) that allow machines to programmatically leverage on their content. 2) create a vectorial representation based on those features for each document and 3) compare them following some distance/divergence functions \cite{Kenter2015}.

In order to reduce the execution time, some approaches have introduced mechanisms (mainly clustering algorithms and pre-election methods) to alleviate the problem of making this calculation over the whole set of pairs in the collection. However those methods are still quite costly.

A novel clustering technique based on topic model distributions is proposed in this paper, in order to reduce the required time to find relations between documents in a large corpus of textual documents without compromising efficiency.

We leverage on Probabilistic Topic Models (PTM) \cite{Blei2010} as representational models and, in particular, Latent Dirichlet Allocation (LDA) \cite{Blei2003} as the way to make this process of finding relations among documents in a corpus more agile and computationally feasible. Probabilistic Topic Modeling techniques \cite{Blei2010a} are statistical methods that analyze the words of the original texts to discover the themes that run through them. Based on these insights, we can further study how those subjects are connected to each other, and how they change over time. Originally developed as a text-mining tool, topic models are now being used to detect instructive structures in data \cite{Blei2012} such as computer vision to classify images \cite{Luo2015}, connect images and captions \cite{Blei2003a}, or build image hierarchies \cite{Bart2011} \cite{Li2010a}; population genetics \cite{Pritchard2000}, and social networks \cite{Jiang2015}. LDA reduces each document to a vector composed by a fixed set of real numbers, each of which represents a probability distribution of a given topic.

One of the main advantages is that PTM's do not require any prior annotations or labeling of the documents. The topics emerge, as hidden structures, from the analysis of the original texts. The \textit{topics} produced by topic modeling techniques are clusters of similar words. A topic model captures this intuition in a mathematical framework, which allows examining a set of documents and discover, based on the statistics of the words contained in each, what the topics might be and what is the topic balance for each document. Those topics offer a much more intuitive, yet sophisticated way of performing knowledge discovery tasks in big collections of documents.

In contrast to existing unsupervised approaches based on centroids or density measures, our algorithm relies on the outcomes of PTM's to assign each document to a cluster without having to consider the other elements in the corpus. Thus, it only takes $O(n)$ time to compute all clusters.


In the following section, we provide an overview of the problem to be solved along with existing solutions. After that, a detailed description of our algorithm is given in Section ~\ref{sec:approach}. We then (Section ~\ref{sec:experiments}) experimentally verify the efficiency and effectiveness of our clustering algorithms using real data, and demonstrate that our approach is competitive enough against both a centroid-based and a density-based clustering baselines. Finally, the most relevant results and conclusions are presented together with some future lines work in Section ~\ref{sec:conclusion}.




\section{Background}
\label{sec:background}

Traditional retrieval tasks over large collections of textual documents \cite{Hearst1999} highly rely on individual features like term frequencies (TF-IDF). However, new ways of characterizing documents based on the automatic generation of models surfacing the main subjects covered in the corpus have been developed during recent years. Probabilistic Topic Modeling \cite{Blei2010} algorithms are statistical methods that analyze the words of the original texts to discover the themes that run through them, how those themes are connected to each other, or how they change over time.

Probabilistic topic models do not require any prior annotations or labeling of the documents. The topics emerge, as hidden structures, from the analysis of the original texts. These structures are topics distributions, per-resource topic distributions or per-resource per-word topic assignments. In turn, a topic is a distribution over terms that is biased around those words associated to a single theme. This interpretable hidden structure annotates each resource in the collection and these annotations can be used to perform deeper analysis about relationships between resources. In this way, topic modeling provides us an algorithmic solution to organize and annotate large collections of textual documents according to their topics.

The simplest generative topic model is \textit{Latent Dirichlet Allocation} (LDA) \cite{Blei2003}. This and other topic models such as \textit{Probabilistic Latent Semantic Analysis} (PLSA) \cite{Hofmann2001} are part of the field known as probabilistic modeling. They are well-known latent variable models for high dimensional data, such as the bag-of-words representation for textual data or any other count-based data representation. While LDA has roots in \textit{Latent Semantic Analysis} (LSA) \cite{Deerwester1990} and PLSA (it was proposed as a generalization of PLSA), it was also influenced by the generative Bayesian framework to avoid some of the over-fitting issues that were observed with PLSA.

This statistical model tries to capture the intuition that documents can exhibit multiple topics. Each document exhibits each topic in different proportion, and each word in each document is drawn from one of the topics, where the selected topic is chosen from the per-document distribution over topics. All the documents in the collection share the same set of topics, but each document exhibits these topics in a different proportion. Documents are  represented as a vector of counts with $W$ components, where $W$ is the number of words in the vocabulary. Each document in the corpus is modeled as a mixture over $K$ topics, and each topic $k$ is a distribution over the vocabulary of $W$ words. Formally, a \textit{topic} is a multinomial distribution over words of a fixed vocabulary representing some concept. Each topic is drawn from a Dirichlet distribution with parameter $\beta$, while each document's mixture is sampled from a Dirichlet distribution with parameter $\alpha$. These two priors, $\alpha$ and $\beta$, are also known as hyper-parameters and they are estimated following some heuristic.

A Dirichlet distribution is a continuous multivariate probability distribution parameterized by a vector of positive reals whose elements sum to 1.  It is \textit{continuous} because the relative likelihood for a random variable to take on a given value is described by a probability density function, and also it is \textit{multivariate} because it has a list of variables with unknown values. In fact, the Dirichlet distribution is the conjugate prior of the categorical distribution and multinomial distribution.

Unlike a restrictive clustering model, where each document is assigned to one cluster, LDA allows documents to exhibit multiple topics. Moreover, since LDA is unsupervised, the topics covered in a set of documents are discovered from the own corpus; the mixed-membership assumptions lead to sharper estimates of word co-occurrence patterns.

\subsection{Similarity Measures Across Documents}
\label{sec:similarity}
In a \textit{Topic Model} the feature vector is a topic distribution expressed as vector of probabilities. Taking into account this premise, the similarity between two topic-based resources will be based on the distance between their topic distributions, which can be also seen as two probability mass functions. A commonly used metric is the \textit{Kullback-Liebler} (KL) divergence. However, it presents two major problems: (1) when a topic distribution is zero, KL divergence is not defined and (2) it is not symmetric, which does not fit well with semantic similarity measures that are usually symmetric \cite{Rus2013}.

\textit{Jensen-Shannon} (JS) divergence \cite{Rao1982}\cite{Lin1991} solves these problems considering the average of the distributions as below \cite{Celikyilmaz2010}:

\begin{equation}
JS(p,q) = \sum\limits_{i=1}^K p_{i}*\log \frac{2*p_{i}}{p_{i}+q_{i}}  +  \sum\limits_{i=1}^K q_{i}*\log \frac{2*q_{i}}{q_{i}+p_{i}}
\label{eq:jsdivergence}
\end{equation}
where  $K$ is the number of topics and $p,q$ are the topics distributions

It can be transformed into a similarity measure as follows \cite{Dagan1998} :

\begin{equation}
sim_{JS}(D_i , D_j) = 10^{- JS(p,q)}
\label{eq:simjs}
\end{equation}
where  $D_i,D_j$ are the documents and $p,q$ the topic distributions of each of them.

\textit{Hellinger} (He) distance is also symmetric and is used along with JS divergence in various fields where a comparison between two probability distributions is required \cite{Blei2007a} \cite{Hall2008} \cite{Boyd-Graber2010}:

\begin{equation}
	He(p, q) = \frac{1}{\sqrt{2}}\cdot\sqrt{\sum\limits_{i=1}^K (\sqrt{p_i} - \sqrt{q_i})^2)}
	\label{eq:hedistance}
\end{equation}

It can be transformed into a similarity measure by subtracting it from 1 \cite{Rus2013} such that a zero distance means max. similarity score and vice versa:

\begin{equation}
	sim_{He}(D_i, D_j) = 1 - He(p,q)
	\label{eq:simhe}
\end{equation}


\section{The Approach}
\label{sec:approach}

Our algorithms draw inspiration from other clustering techniques to divide the initial space of elements into smaller sub-groups where the complexity of calculating all possible distances is significantly reduced. Existing unsupervised approaches based on centroids or density measures require to make comparisons between elements to find groups of similar elements in the collection. They normally follow an iterative methodology to produce the final solution, based on calculating distances between the elements inside each intermediate state. A na{\"i}ve approach would need to calculate all possible distances between elements, which takes $O(n^2)$ time for a $n \times n$ matrix. That makes it impossible to apply such techniques on large collections of documents, since the cost of comparing each element with the others escalates quickly. For those big volumes of data, a clustering task that only takes linear time to discover the clusters can significantly alleviate this problem. For example, a classification method that does not require any other data except the element information to assign the item to the corresponding cluster will take $O(n)$ time to compose those groups.

The classification method needs to take advantage of both the vectorial representations of the documents and the similarity measure used to relate them in a corpus. Since the representational model considered is based on Probabilistic Topic Models (and more specifically on LDA), the classification method leverages on the particular behavior of Dirichlet distributions, which describes each document by a density vector where the sum of all the probability values must be equal to 1.0. Thus, analyzing the relations between the topics that compose a topic distribution becomes more important than comparing their probability values with another topic distribution.

Our hypothesis is that, given a collection of topic distributions, an unsupervised classification with high precision and linear computing time can be performed by considering only the topic distribution of each document and without needing to further compare it with other document's distributions.

All algorithms have been compared in terms of \textit{cost}, \textit{effectiveness} and \textit{efficiency} \cite{Halkidi2001a}. \textit{Cost} is based on the number of pairwise similarity values. \textit{Effectiveness} handles relevance measures such as \textit{precision} and \textit{recall}. And \textit{efficiency} tries to measure the overall balance between \textit{cost} and \textit{effectiveness}. More details about those measures will be included in Section~\ref{sec:experiments}.

\subsection{Trends-based Clustering}
Topic distributions are formalized as probability distributions following a Dirichlet distribution, so their probability values sum to 1. In this way, the relevance of a topic is influenced and at the same time influences the relevance of the others items in the distribution. Our first approach named \textit{Trends on Dirichlet distribution-based Clustering} (TDC) considers changes in the relevance, i.e. probability values of the topics instead of directly relying on the scores associated to a given topic distribution. It expresses the oscillations between topic weights considering a fixed order between them. The order can be any, as long as it remains constant in all distributions. Thus, a \textit{probability-vector} composed by $n$ density values is translated to a \textit{trend-expression} made out of $n-1$ trend-values such as (1) upward, (2) downward and (0) sustained. This \textit{trend-expression} will identify the cluster  the distribution falls into, and therefore the corresponding item belongs to. TDC is defined as:
\begin{equation}
TDC(P)=T
\end{equation}
\begin{conditions}
 T_{i}     & 1,  when $P_i < P_{i+1}$ \\
 T_{i}     & 2,  when $P_i > P_{i+1}$ \\   
 T_{i} 	   & 0,  when $P_i = P_{i+1}$
\end{conditions}
For example, given the distribution $P_1=[0.23, 0.18, 0.33, 0.13, 0.13]$, the assigned cluster will be $T=2120$. The first value is $2$ because $0.23$ is greater than $0.18$ (same for other values).

\subsection{Ranking-based Clustering}
We propose a clustering technique named \textit{Ranking on Dirichlet distribution-based Clustering} (RDC) that only considers the top $n$ topics from the ranked list of probability distributions to classify similar topic distributions. It is based on the focal document selection proposed by \cite{Towne2016} to validate LDA-based similarity algorithms against human perception of similarity. RDC is defined as:
\begin{equation}
RDC(P)=R
\end{equation}
where  $\forall i \in R, R_i>=R_{i+1}$ and $\forall j \in P, R_1>=P_j$

This is based on the assumption that the highest weighted topics have a high influence in the rest of topics in terms of calculating distances, when comparing continuous multivariate probability distributions. Since similarity measures (Section ~\ref{sec:similarity}) based on probability distributions are oriented to determine the uncertainty of the distribution, when a mixture of probability distributions is considered, as in the case of Topic Models, the top $n$ distributions (i.e. the most relevant topics) should be sufficient to allow us grouping similar distributions. Taking into account the above considerations, the RDC algorithm classifies a topic distribution according to only $n$ highest probability values. For instance, given the following topic distribution: $P_2=[0.23, 0.18, 0.33, 0.13, 0.13]$, the assigned cluster is \textit{3} from RDC-1 because that is the topic with the highest weight.

\subsection{Cumulative Ranking-based Clustering}
A variant of the previous algorithm, named \textit{Cumulative Ranking on Dirichlet distribution-based Clustering} (CRDC), also aims to discover the most representative topics that can help to group similar topic distributions. While RDC is based on a fixed number of topics, CRDC is based on the cumulative sum of the weights of the highest topics. The number of topics is now dynamically determined by a threshold, and once this threshold is reached no more topics are considered. CRDC is defined as:
\begin{equation}
CRDC(P)=C
\end{equation}
where  $\forall i \in C, C_i >=C_{i+1}$ and $\sum\limits_{i=1}^T C_i >= w$ with $T$ size of $C$, and $w$ a cumulative weight threshold.

For instance, considering a CRDC algorithm considering a cumulative weight threshold of 0.9, and the following topic distribution: $P_3=[0.36, 0.58, 0.05, 0.01]$. The assigned cluster will be \textit{2|1}. To come up with this cluster, a ranked list of topics based on their weights is first calculated, $R_p=2|1|3|4$. Then, a sum of weights according to the order described by $R_p$ is performed. When the accumulated sum is greater than the threshold, the topics taking part of the sum will be selected to ``label'' the cluster. In this case, the cumulative weight threshold is 0.9 therefore using only the first two topics we exceed the threshold: $w=0.58+0.36=0.94$

\section{Experiments}
\label{sec:experiments}
In this section we present the experimental setup for evaluating our trends-based (TDC), ranking-based (RDC) and cumulative ranking-based (CRDC) clustering approaches, considering both JS divergence and He distance as similarity measures. We describe the datasets and baseline algorithms that will be used for comparison.

\subsection{Datasets}
\label{sec:datasets}
We used two datasets to evaluate the performance of the algorithms. The first dataset, DIRICHLET-RANDOM-MIXTURE (DRM), is synthetic \cite{Badenes-Olmedo2017a}. To generate the dataset, we sampled \textit{k} probabilistic distributions from a \textit{randomly k-dimensional selector} based on Dirichlet distributions. This implies that all probabilities must to sum to 1 for each sampled point. The number of sampled points from this mixture of Dirichlet distributions is $n=1000$.

The second dataset has been created from a collection of research papers published in the \textit{Advances in Engineering Software} (AIES) journal. They were retrieved from the Springer API by using the librAIry \cite{Badenes-Olmedo2017} framework and a Topic Model based on LDA was created from them. The sample is also composed by $n=1000$ documents.

Topic models were trained from these datasets by using the criteria described by \cite{Steyvers2006}: $\alpha= 50/k$ , $\beta= 0.01$ and $k=2*(\sqrt{(n/2)})$, where $k$ is the number of topics and $n$ is the number of documents. Since both datasets contain $1000$ documents ($n$), the hyper-parameters $\alpha$ and $\beta$ are assigned as follow: $\alpha=1.136$, and $\beta=0.01$, and the number of topics is fixed to $k=44$. Further tuning of the settings is not crucial in this evaluation process, because we are not focusing on the quality of the model but on the efficiency when calculating similarities from their representational distributions.

\subsection{Similarity Threshold}
\label{sec:threshold}

\begin{figure}
  \includegraphics[scale=0.33]{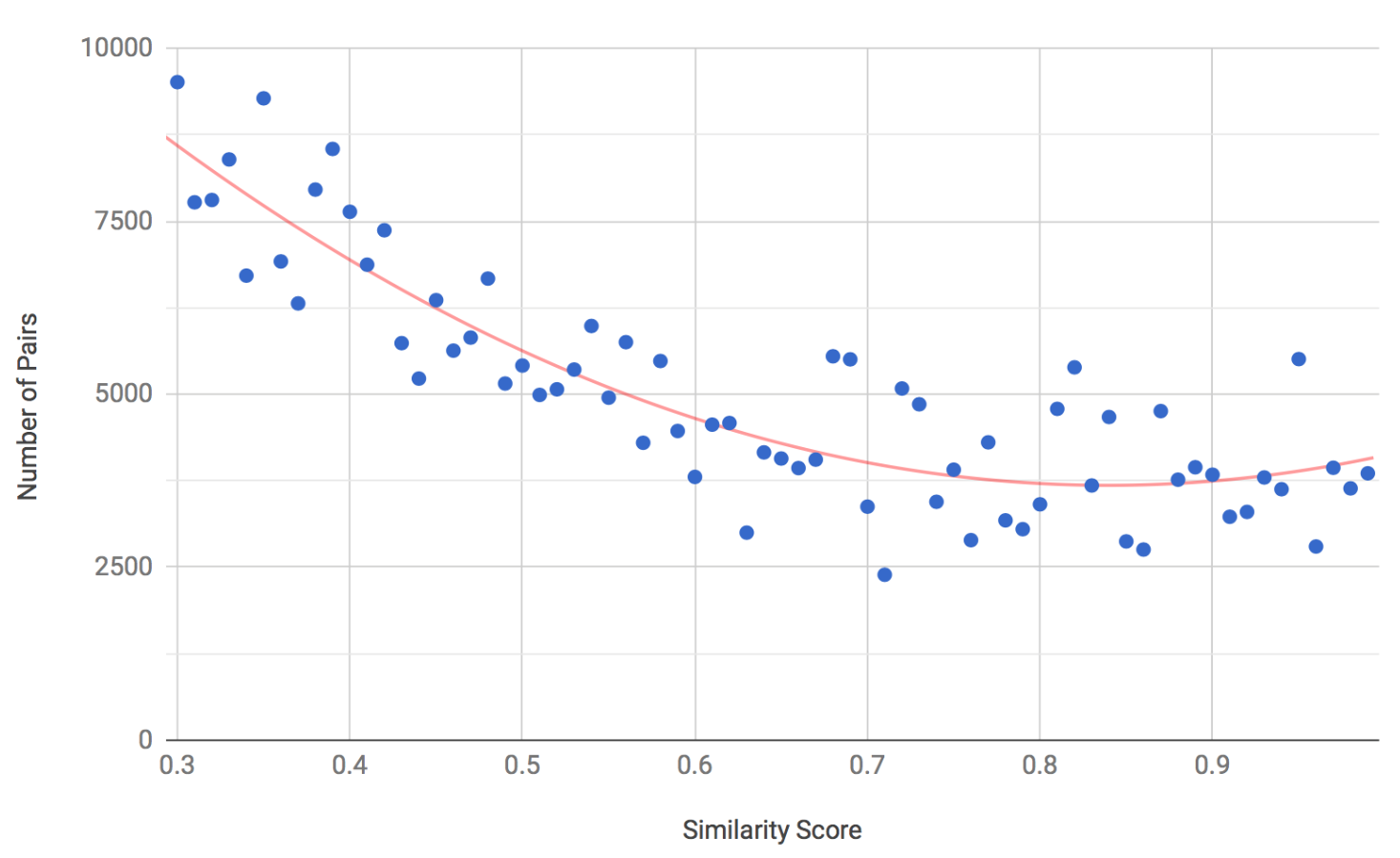}
  \caption{Similarity values grouped by frequency in AIES}
  \label{fig:similarityThreshold}
\end{figure}

Since there is no unified criteria to select a threshold inside the distance scores spectrum that allows us to determine when two documents are similar, we decided to study the distribution of similarity values calculated from all pairwise comparisons. In Figure ~\ref{fig:similarityThreshold}, the result of grouping all similarities by the two most representative decimals, i.e. the first two decimals of the similarity value, is shown. Then, a polynomial function (red line) is approximated to describe the trend of these values. In this function, the similarity score $0.83$ emerges as a global minimum and has been used for filtering out the non-similar document pairs.

\subsection{Baselines}
We compare the performance of TDC, RDC and CRDC algorithms against the following baselines:
\begin{itemize}
  \item \textit{K-Means} as a centroid-based clustering approach.
  \item \textit{DBSCAN} as a density-based clustering approach.
  \item \textit{Random}, which randomly selects $R$ from the dataset
\end{itemize}

Initially, \textit{K-Means} \cite{Bahmani2012} randomly composes a set of centroids and assigns each point of the sample to its nearest cluster based on a distance measure. Then, a new set of centroids is calculated from the previous ones according to the assigned points. This process is repeated until the set of centroids does not change significantly between consecutive iterations or a maximum number of iterations is reached. The \textit{scalable K-Means} approach used in our experiments is an improved version of \textit{k-means} which obtains an initial set of centers ideally close to the optimum solution. The algorithm implemented at the Apache Commons Math library \footnote{ http://commons.apache.org/proper/commons-math/} was used in the experiments. Based on empirical results, the best configuration is: $k=number-of-topics=44$ and $maxIterations=50$

A widely known density-based algorithm is \textit{DBSCAN} \cite{Ester1996}, which compose clusters from the neighborhood of each point considering at least a minimum number of points and a given radius. Thus, it requires to specify the radius of the point's neighborhood, \textit{Eps}, and the minimum number of points in the neighborhood \textit{MinPts}. Based on empirical results, the best results were obtained with the following configuration: $eps=0.1$ and $minPts=50$

The \textit{Random} algorithm takes as input a parameter \textit{m} and randomly divides the dataset into \textit{m} equal-sized groups of similar documents. For the evaluation, $m$ was set to the number of topics, the dimension of the dataset.

With respect to the proposed algorithms and taking into account empirical results, the RDC algorithm is set to use the \textit{top1} highest topics, and the cumulative weight threshold for the CRDC algorithm is set to $0.9$.

\subsection{Measure}
A gold-standard is created for each dataset and distance metric considered. They are created by calculating all pairwise similarities from their documents. Since the $n \times n$ similarity matrix requires $O(n^2)$ time to be calculated, the selected size of datasets has not been too large $n=1000$.

We considered three measures to evaluate our algorithms with respect to the baseline:
\begin{itemize}
  \item \textbf{\textit{cost}}: based on the number of similarity score calculations required by the algorithm:
\begin{equation}
cost=(reqSim - minSim)/(totalSim - minSim)
\end{equation}
The \textit{minSim} corresponds to the number of similar documents obtained from using the \textit{threshold} score previously mentioned in section ~\ref{sec:threshold}. The \textit{totalSim} corresponds to the Cartesian product of existing documents: $totalSim=n*n=1,000,000$. And the \textit{reqSim} corresponds to the number of similarities calculated by the algorithm.
  \item \textbf{\textit{effectiveness}}: based on $precision$ and $recall$. It expresses the quality of the algorithm:
\begin{equation}
 effectiveness = \frac{precision^2  + recall^2}{2}
\end{equation}
  \item \textbf{\textit{efficiency}}: based on the previous ones, it express a compromise between quality and performance:
\begin{equation}
 efficiency = effectiveness - cost
\end{equation}
\end{itemize}

\subsection{Results}

The code used to evaluate the algorithms along with the results obtained are available on GitHub \cite{Badenes-Olmedo2017a}.

\begin{figure}
  \includegraphics[scale=0.27]{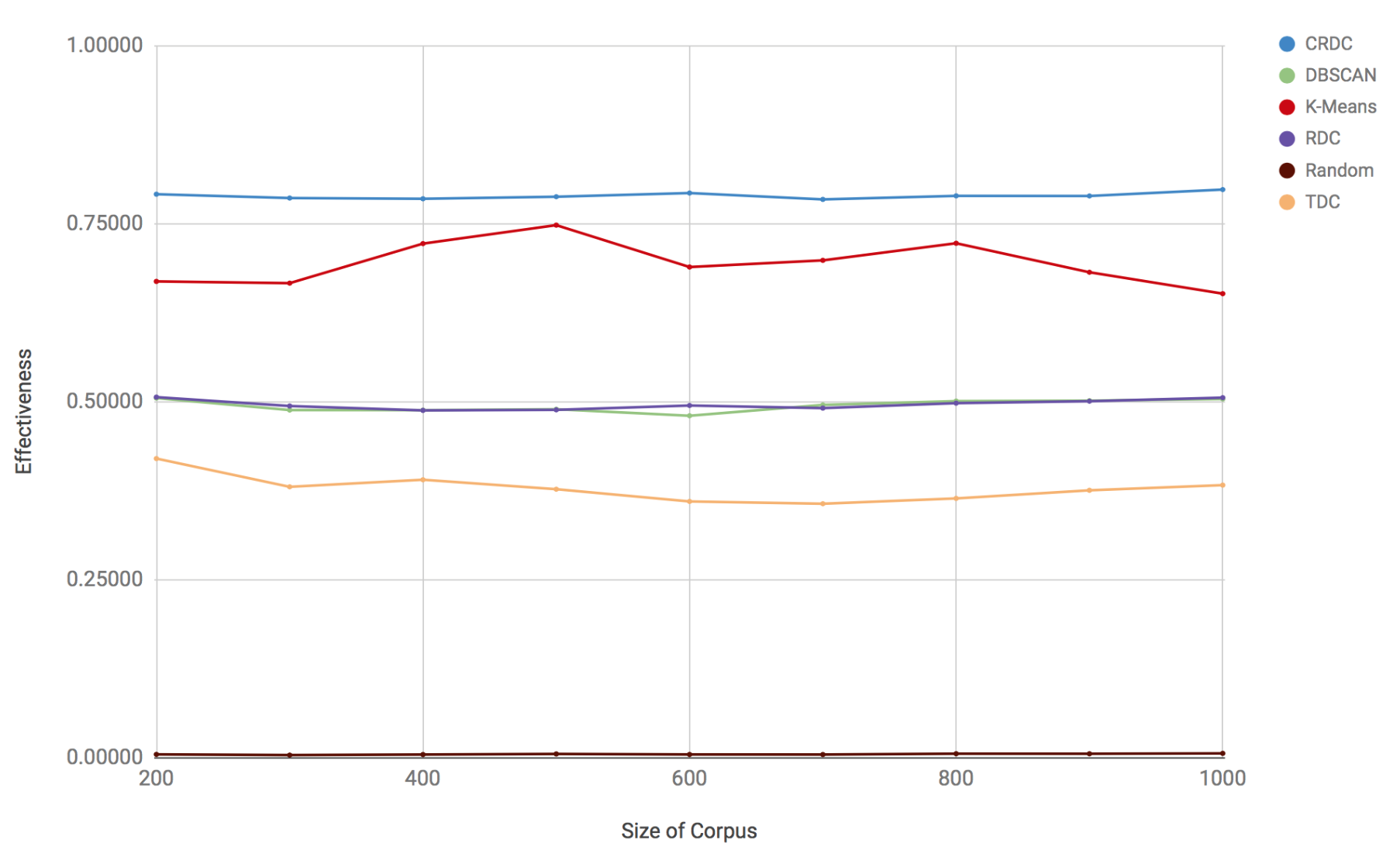}
  \caption{Effectiveness (JS-based) in AIES}
  \label{fig:effectivenessJS}
\end{figure}

\begin{figure}
  \includegraphics[scale=0.27]{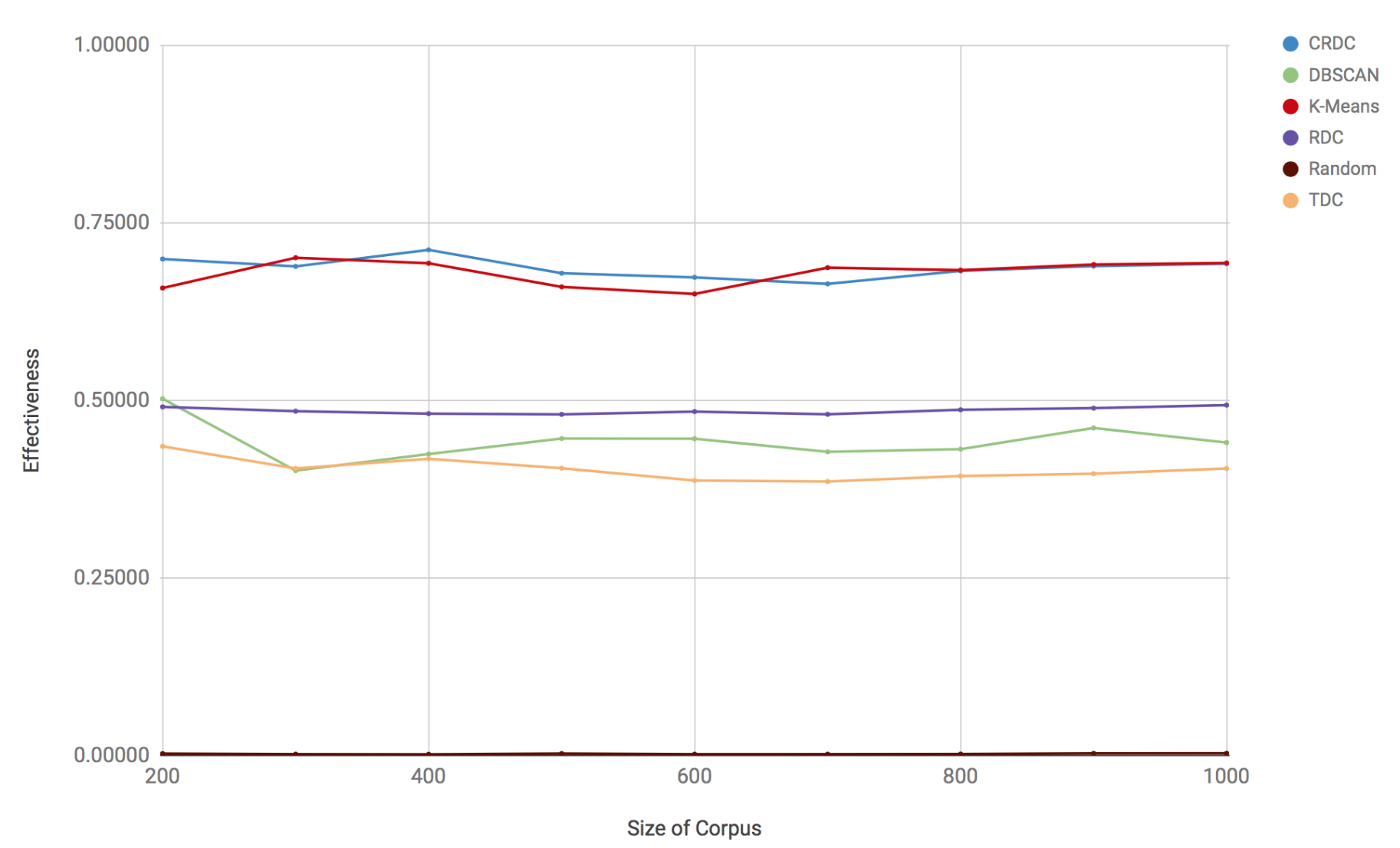}
  \caption{Effectiveness (He-based) in AIES}
  \label{fig:effectivenessHe}
\end{figure}

\begin{figure}
  \includegraphics[scale=0.27]{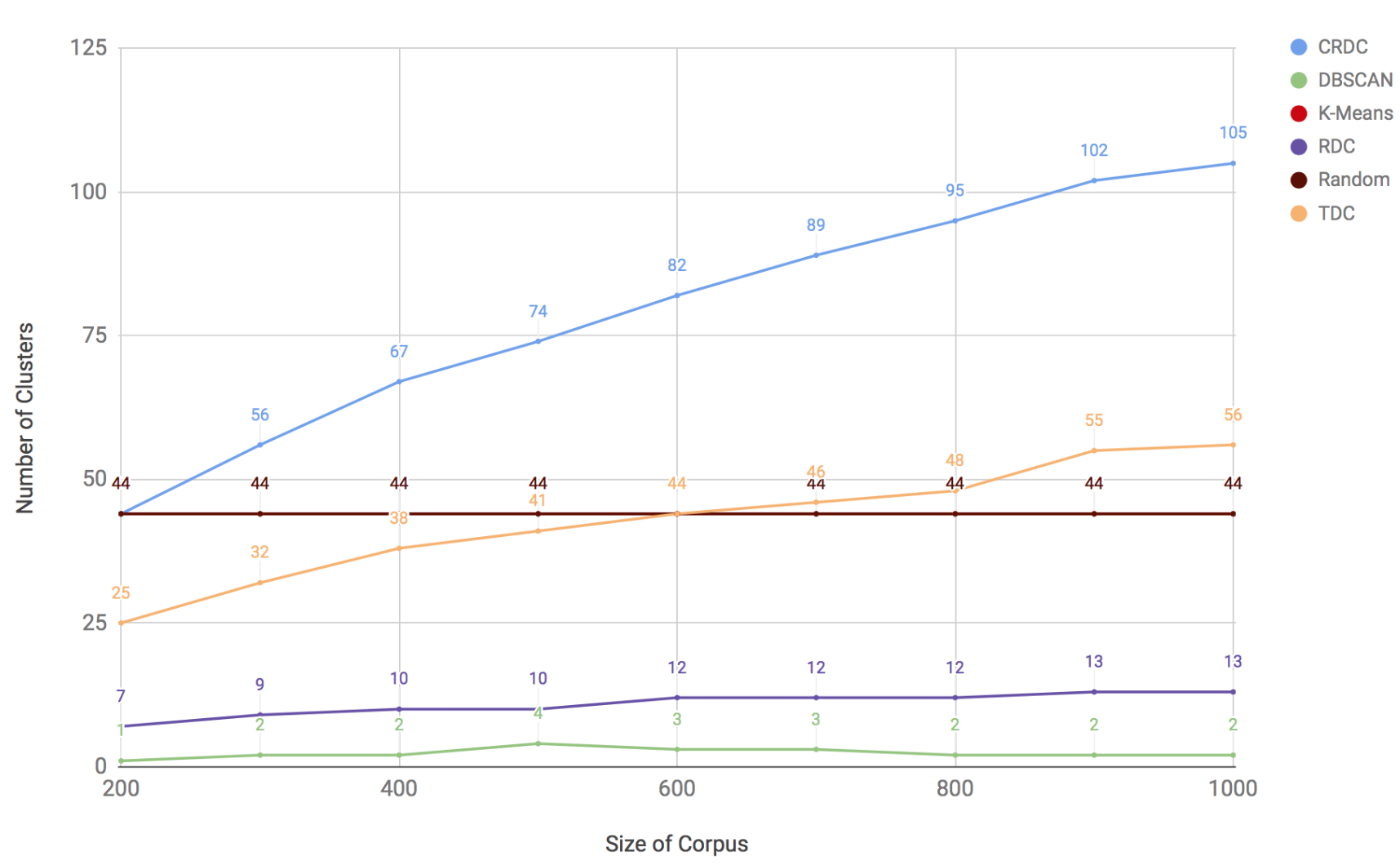}
  \caption{Clusters in AIES}
  \label{fig:clusters}
\end{figure}

\textbf{In terms of \textit{effectiveness}} (Figures ~\ref{fig:effectivenessJS} and ~\ref{fig:effectivenessHe}), the results highlight that \textit{K-Means} and \textit{CRDC} outperform the other algorithms. \textit{K-Means} was expected to be a top performer because the algorithm itself performs comparisons to map clusters. The fact that \textit{CRDC} has such good performance encourages us to think that, in fact, the most relevant topics when they altogether exceed a certain high weight threshold, are those that best represent the document and allow to group together similar documents. However, as shown in tables ~\ref{tab:precisionJS}, ~\ref{tab:precisionHe}, ~\ref{tab:recallJS} and ~\ref{tab:recallHe}, considering a fixed number of more relevant topics (\textit{RDC}) or considering the trend of their weights (\textit{TDC}) does not seem to perform so well on aggregating similar documents, since their \textit{precision} and \textit{recall} values are very low in both cases. It is surprising that the \textit{DBSCAN} has such low value. Taking a look at its \textit{precision} and \textit{recall} values, and also seeing the number of groups that each algorithm has created (Figure ~\ref{fig:clusters}), we believe that having a corpus containing a very cohesive set of documents (all papers in corpus belong to the same journal) affects the performance of this algorithm since it divides the corpus into a lower number of groups. This way, it obtains high values of \textit{recall} because most of the pair-wise distances are computed, but very low \textit{precision}.

The results also show that the behavior of the algorithms does not differ significantly when using different similarity measures, for example JS divergence (Figure ~\ref{fig:effectivenessJS}) and He distance (Figure ~\ref{fig:effectivenessHe}). This highlights the importance of the documents' topic distributions to successfully classify them into smaller groups of similar items, while other particular aspects such as the distance or similarity metric used to compare them are less influential.

\begin{table}[!htb]
    \centering
        \begin{tabular}{l*{6}{c}r}\hline
                    Size    & CRDC & DBSCAN & K-Means & RDC & TDC  & Random \\
          \hline
					200 & 0.94 & 0.10 & \textbf{0.96} & 0.31 & 0.42 & 0.12 \\
					300 & 0.93 & 0.15 & \textbf{0.94} & 0.30 & 0.39 & 0.08 \\
					400 & \textbf{0.93} & 0.15 & 0.89 & 0.29 & 0.39 & 0.09 \\
					500 & \textbf{0.92} & 0.30 & 0.90 & 0.28 & 0.38 & 0.09 \\
					600 & \textbf{0.92} & 0.19 & 0.88 & 0.28 & 0.38 & 0.08 \\
					700 & \textbf{0.92} & 0.20 & 0.91 & 0.28 & 0.38 & 0.09 \\
					800 & \textbf{0.92} & 0.12 & 0.89 & 0.30 & 0.39 & 0.10 \\
					900 & \textbf{0.92} & 0.13 & 0.87 & 0.30 & 0.40 & 0.10 \\
					1000 & \textbf{0.93} & 0.13 & 0.90 & 0.30 & 0.40 & 0.10 \\
        \end{tabular}
    \caption{Precision (JS-based) in AIES}\label{tab:precisionJS}
\end{table}

\begin{table}[!htb]
    \centering
        \begin{tabular}{l*{6}{c}r}\hline
                    Size  & CRDC & DBSCAN & K-Means & RDC & TDC  & Random \\
          \hline
					200 & 0.75 & 0.07 & \textbf{0.84} & 0.23 & 0.08 & 0.33 \\
					300 & 0.74 & 0.08 & \textbf{0.83} & 0.23 & 0.06 & 0.32 \\
					400 & \textbf{0.76} & \textbf{0.09} & 0.76 & 0.22 & 0.06 & 0.32 \\
					500 & 0.73 & 0.08 & \textbf{0.74} & 0.21 & 0.08 & 0.31 \\
					600 & 0.72 & 0.08 & \textbf{0.73} & 0.21 & 0.06 & 0.30 \\
					700 & 0.71 & 0.10 & \textbf{0.76} & 0.21 & 0.06 & 0.30 \\
					800 & 0.73 & 0.11 & \textbf{0.78} & 0.22 & 0.07 & 0.31 \\
					900 & 0.73 & 0.12 & \textbf{0.80} & 0.22 & 0.08 & 0.32 \\
					1000 & 0.74 & 0.15 & \textbf{0.77} & 0.23 & 0.08 & 0.32 \\
        \end{tabular}
    \caption{Precision (He-based) in AIES}\label{tab:precisionHe}
\end{table}

\begin{table}[!htb]
    \centering
        \begin{tabular}{l*{6}{c}r}\hline
                    Size  & CRDC & DBSCAN & K-Means & RDC & TDC  & Random \\
          \hline
					200  & 0.92  & \textbf{1.00}  & 0.79  & 0.96  & 0.02  & 0.87 \\
					300  & 0.91  & 0.89  & 0.84  & \textbf{0.96}  & 0.02  & 0.84 \\
					400  & 0.92  & 0.92  & 0.90  & \textbf{0.96}  & 0.02  & 0.86 \\
					500  & 0.91  & 0.94  & 0.88  & \textbf{0.96}  & 0.03  & 0.85 \\
					600  & 0.91  & 0.94  & 0.87  & \textbf{0.96}  & 0.02  & 0.83 \\
					700  & 0.91  & 0.92  & 0.90  & \textbf{0.96}  & 0.02  & 0.83 \\
					800  & 0.92  & 0.92  & 0.88  & \textbf{0.96}  & 0.02  & 0.83 \\
					900  & 0.92  & 0.95  & 0.86  & \textbf{0.96}  & 0.02  & 0.83 \\
					1000  & 0.92  & 0.93  & 0.89  & \textbf{0.97}  & 0.02  & 0.84 \\
        \end{tabular}
    \caption{Recall (JS-based) in AIES}\label{tab:recallHe}
\end{table}

\begin{table}[!htb]
    \centering
        \begin{tabular}{l*{6}{c}r}\hline
                    Size`  & CRDC & DBSCAN & K-Means & RDC & TDC  & Random \\
          \hline
					200 & 0.84 & \textbf{1.00} & 0.65 & 0.96 & 0.02 & 0.82 \\
					300 & 0.84 & \textbf{0.98} & 0.76 & 0.95 & 0.02 & 0.78 \\
					400 & 0.84 & \textbf{0.98} & 0.79 & 0.94 & 0.02 & 0.79 \\
					500 & 0.85 & 0.94 & 0.87 & \textbf{0.95} & 0.02 & 0.78 \\
					600 & 0.86 & \textbf{0.96} & 0.80 & 0.95 & 0.02 & 0.76 \\
					700 & 0.85 & \textbf{0.98} & 0.80 & 0.95 & 0.02 & 0.76 \\
					800 & 0.85 & \textbf{0.99} & 0.81 & 0.95 & 0.02 & 0.76 \\
					900 & 0.85 & \textbf{0.99} & 0.75 & 0.95 & 0.02 & 0.77 \\
					1000 & 0.86 & \textbf{1.00} & 0.74 & 0.96 & 0.02 & 0.78 \\
        \end{tabular}
    \caption{Recall (He-based) in AIES}\label{tab:recallJS}
\end{table}

\begin{figure}
  \includegraphics[scale=0.27]{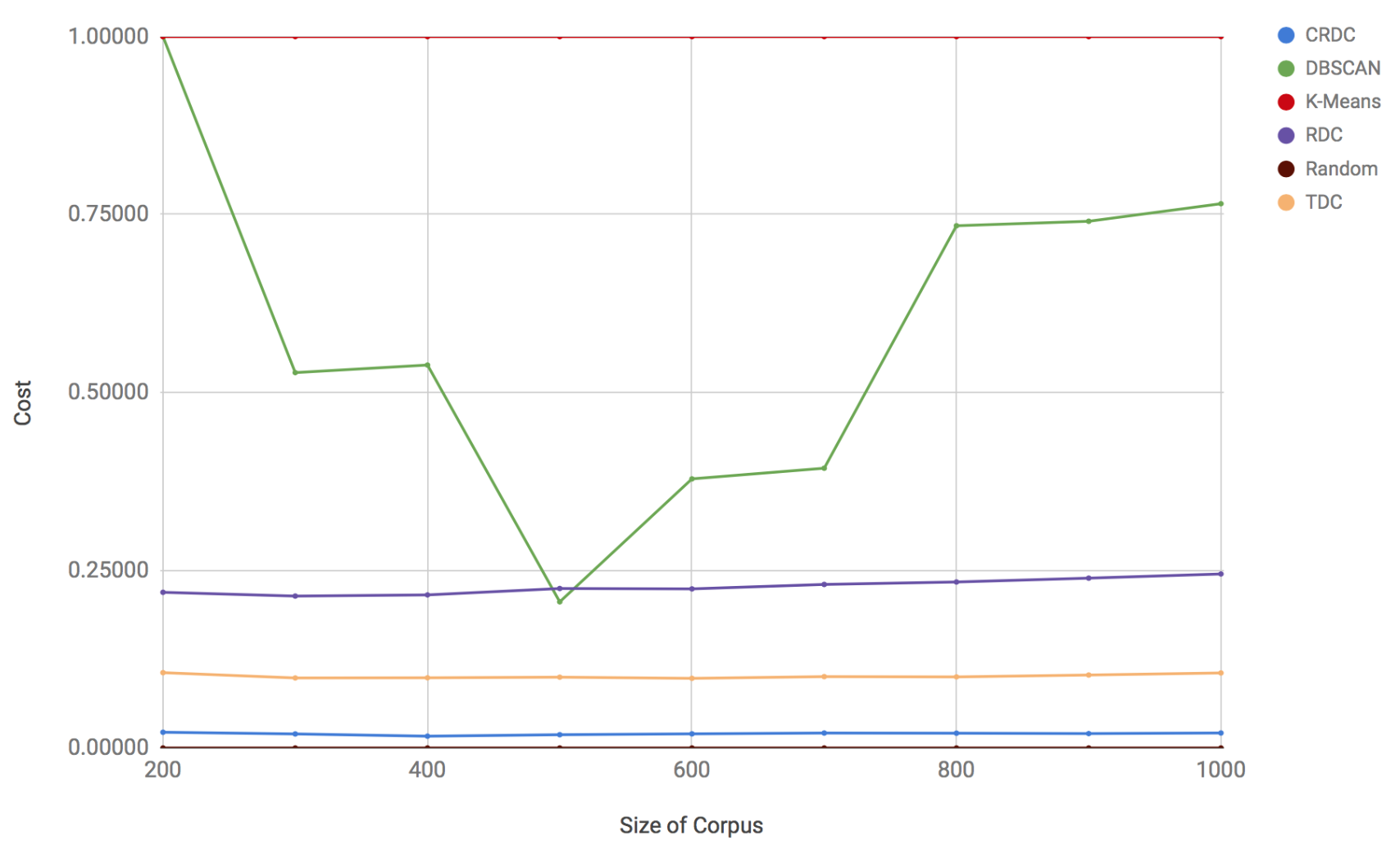}
  \caption{Cost (JS-based) in AIES}
  \label{fig:costJS}
\end{figure}

\begin{figure}
  \includegraphics[scale=0.27]{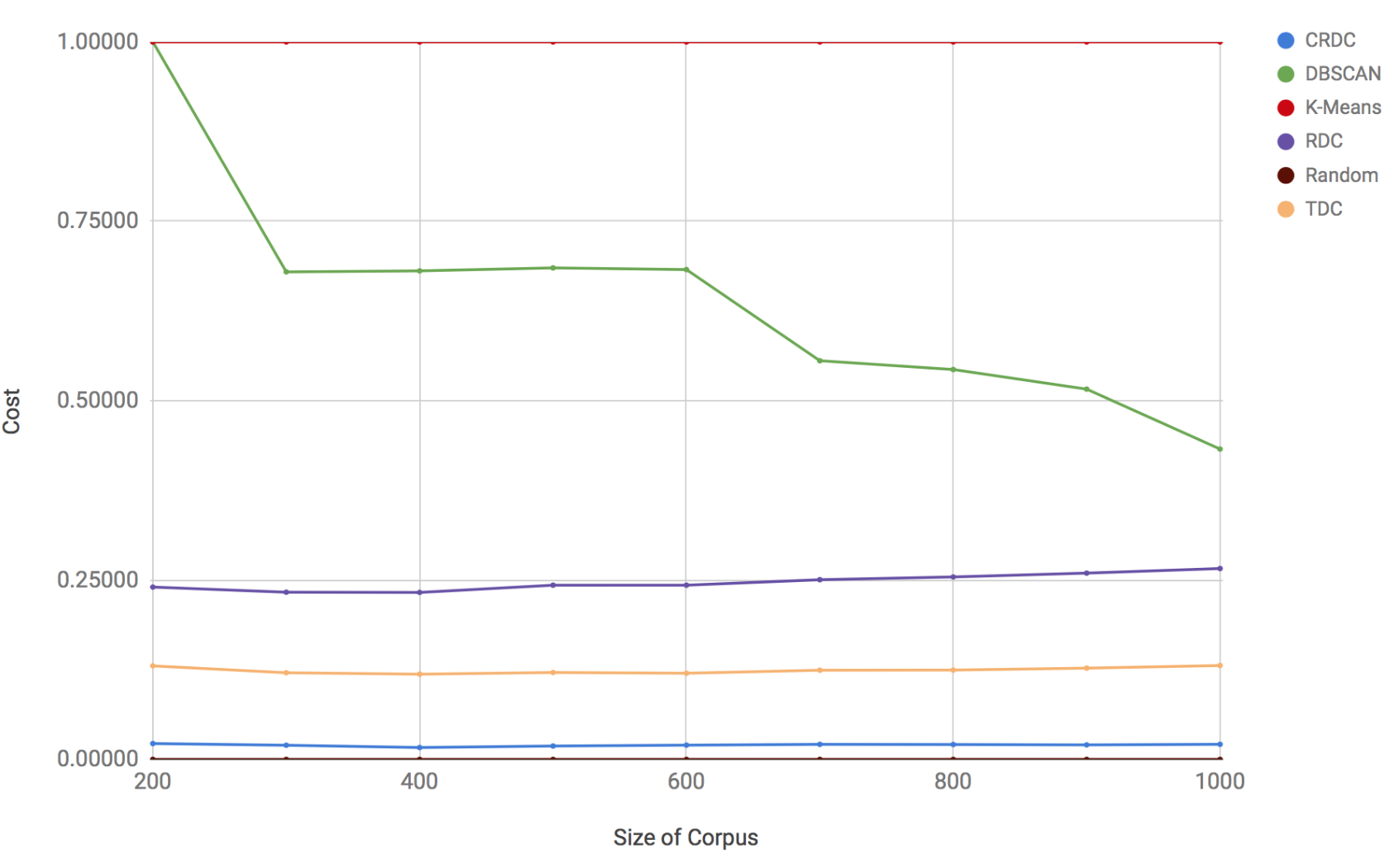}
  \caption{Cost (He-based) in AIES}
  \label{fig:costHe}
\end{figure}

\textbf{In terms of \textit{cost}} (Figures ~\ref{fig:costJS} and ~\ref{fig:costHe}), the best clustering algorithm, as expected, is based on \textit{random} selection. This is due to the fact that the number of pairs compared by this algorithm is always the minimum, given the dataset is simply randomly divided into \textit{m} equal-sized groups, where \textit{m} is equals to the number of topics, i.e. dimension of the dataset. Since \textit{K-Means} and \textit{DBSCAN} make comparisons between documents until their internal condition is satisfied, they are the most inefficient approaches. \textit{K-Means} involves the highest cost because it compares all the documents with the 44 centroids in each iteration.

Among our proposals, the main reason for an algorithm to present a higher cost is due to the number of groups the corpus is divided into (see Figure ~\ref{fig:clusters}). The greater the number of groups, the fewer the number of later comparisons that have to be made and, therefore, the lower the cost of the algorithm.

The behavior of the \textit{DBSCAN} algorithm depends remarkably on the similarity metric used. We think that this may be due to the way in which both measures satisfy the triangle inequality condition, since one is based on divergence (JS) and the other on distance (He). This property, which defines $distance(a,b) \leq distance(a,c) + distance(c,b)$, is very important in the calculations that \textit{DBSCAN} makes to discover the groups, since it only calculates the distances between near points.


\begin{figure}
  \includegraphics[scale=0.27]{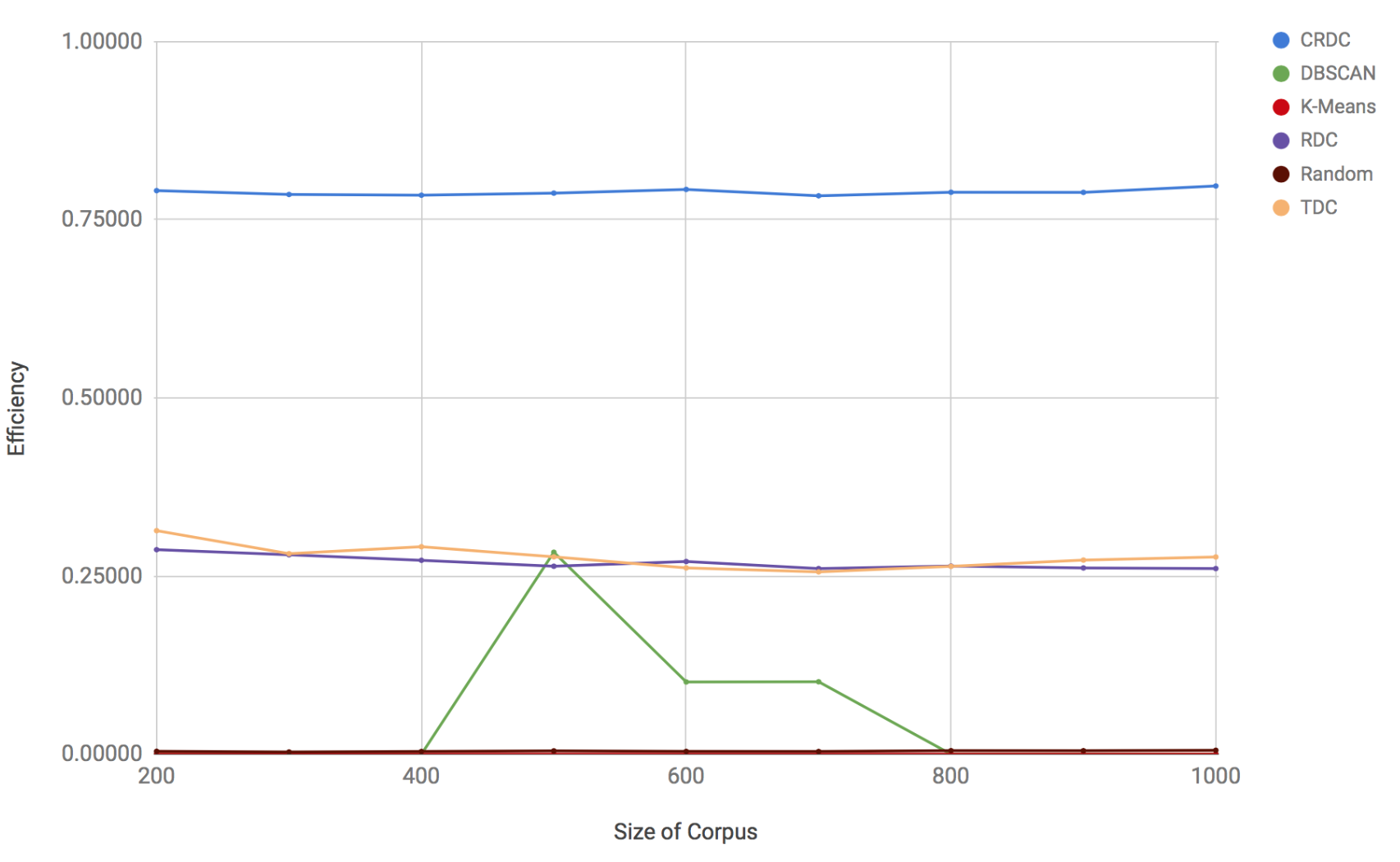}
  \caption{Efficiency (JS-based) in AIES}
  \label{fig:efficiencyJS}
\end{figure}

\begin{figure}
  \includegraphics[scale=0.27]{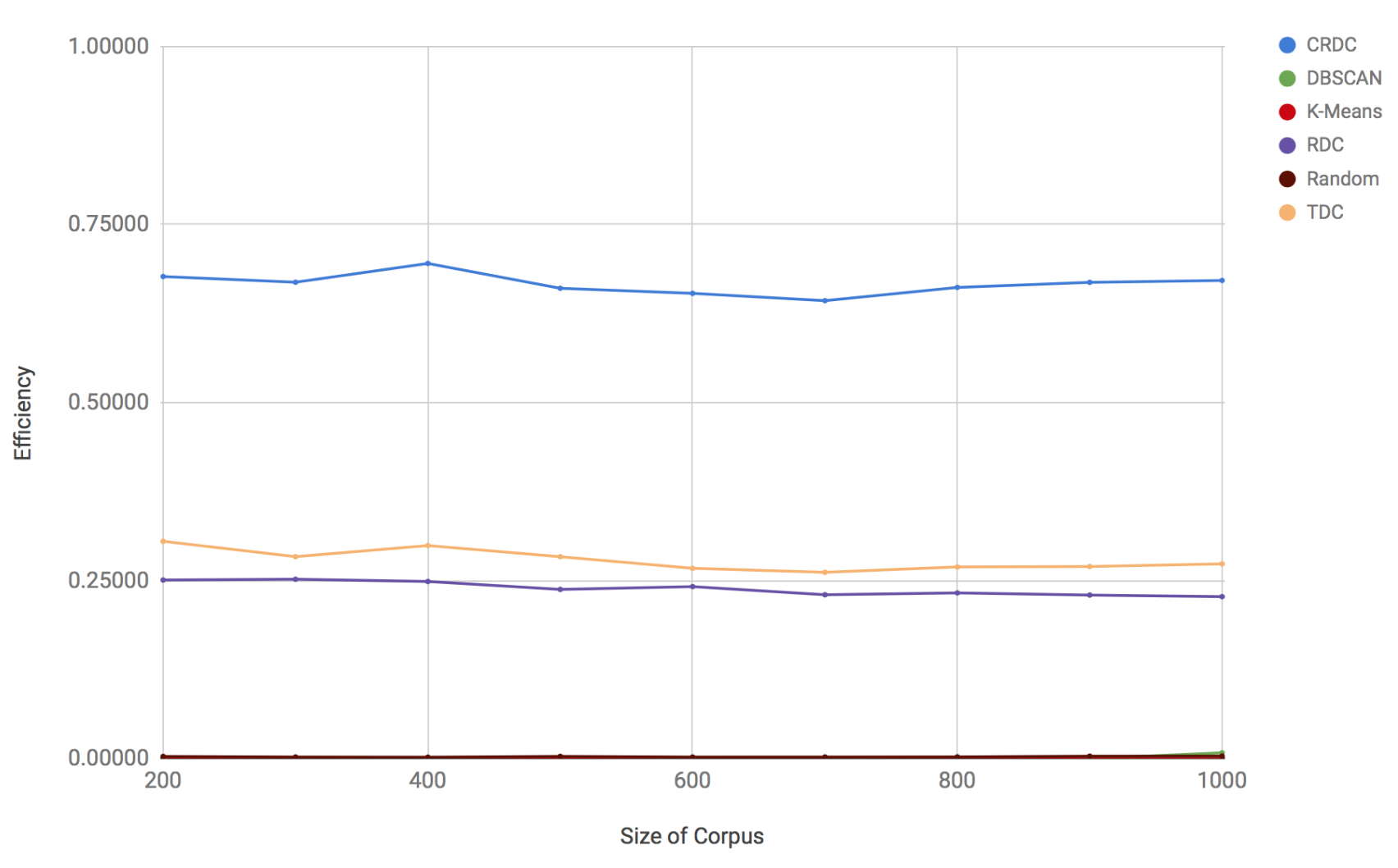}
  \caption{Efficiency (He-based) in AIES}
  \label{fig:efficiencyHe}
\end{figure}

Finally, \textbf{in terms of \textit{efficiency}} (Figures ~\ref{fig:efficiencyJS}, ~\ref{fig:efficiencyHe}), regardless of the similarity measure used, the algorithm that yields the best performance according to the results obtained is \textit{CRDC}. Overall, \textit{CRDC} demonstrates a high accuracy classification and a lower cost by improving the performance offered by centroid-based or density-based approaches.

\begin{figure}
  \includegraphics[scale=0.30]{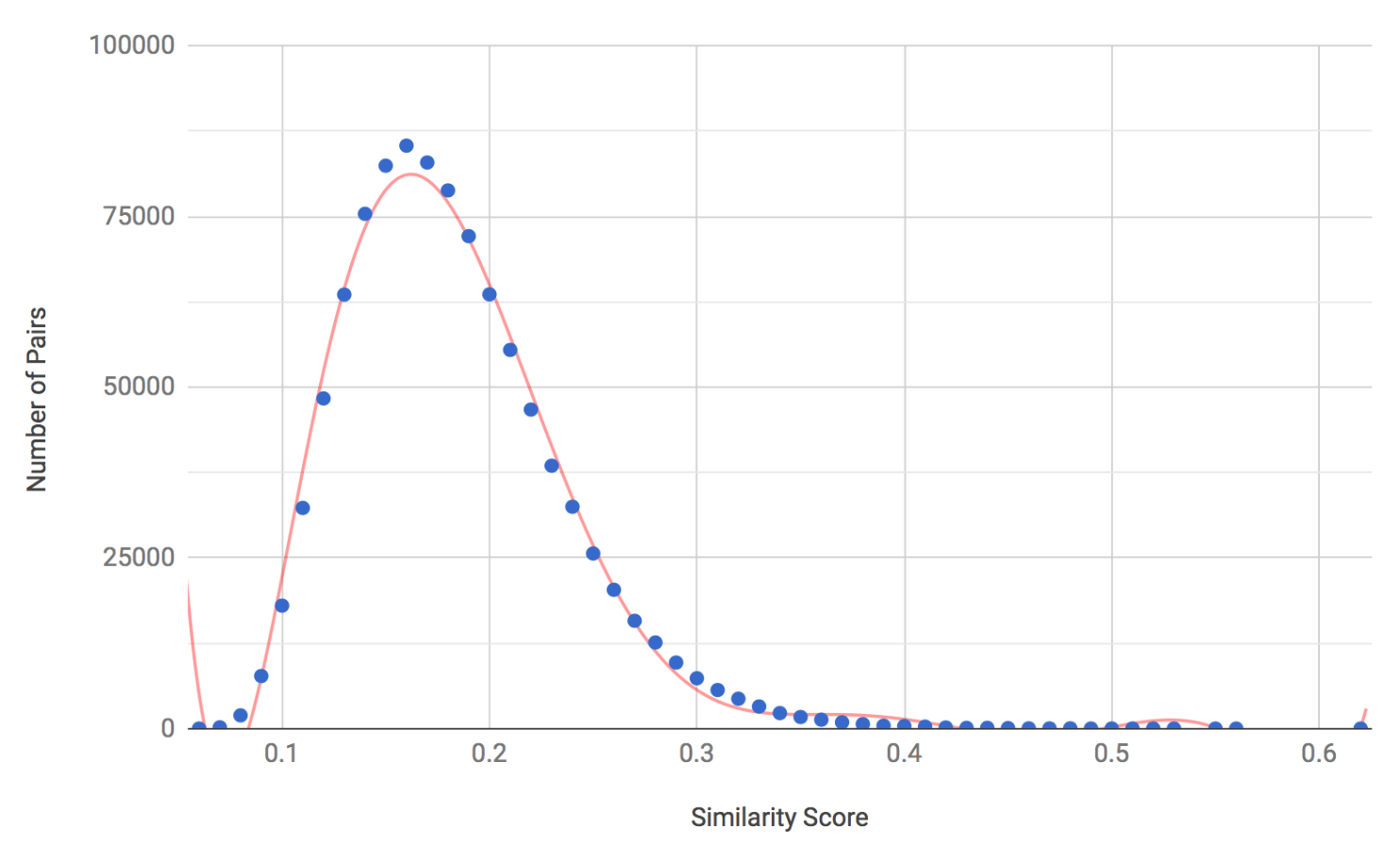}
  \caption{Similarity values grouped by frequency in DRM}
  \label{fig:similaritiesDRM}
\end{figure}

\begin{figure}
  \includegraphics[scale=0.27]{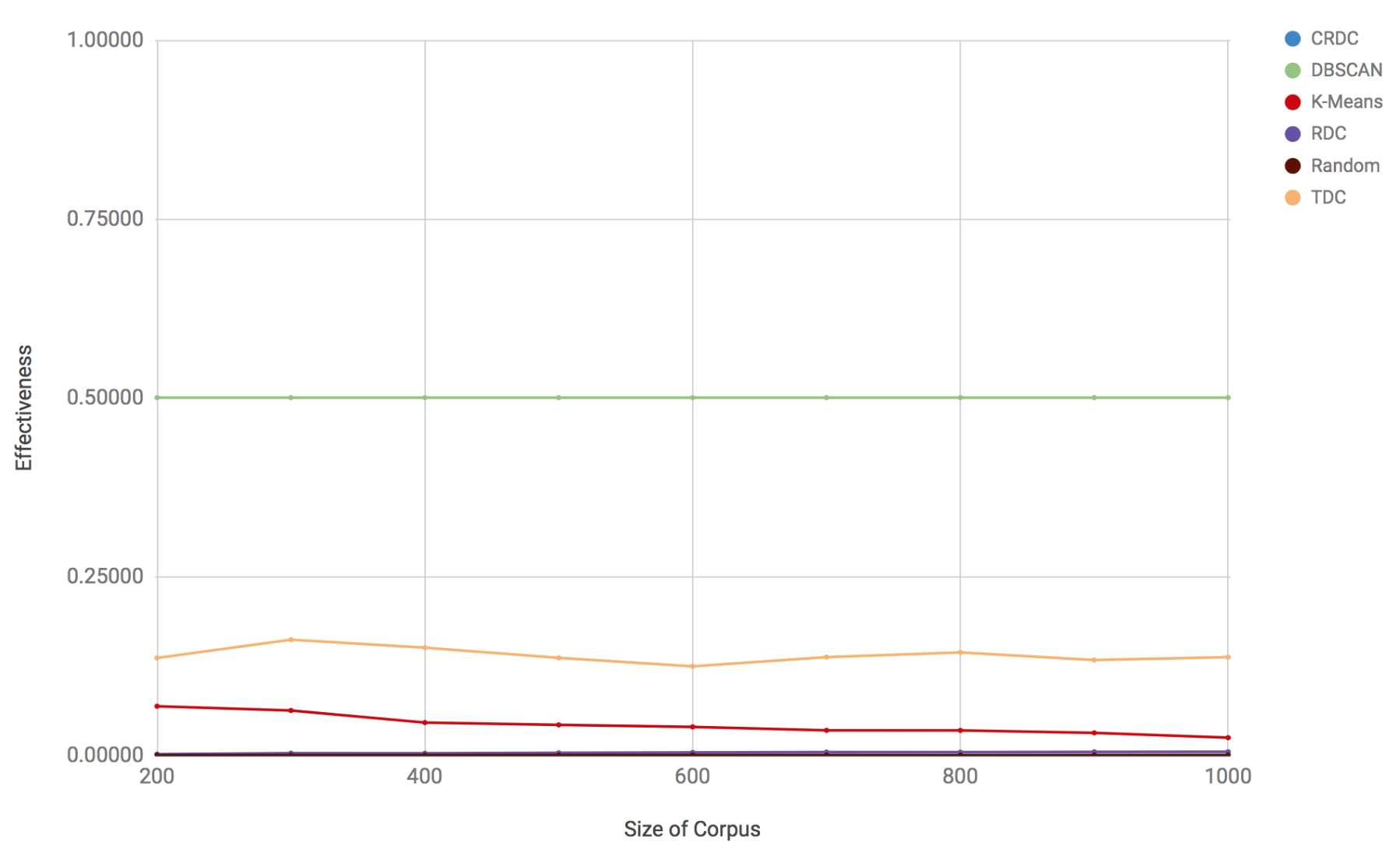}
  \caption{Effectiveness (JS based) in DRM}
  \label{fig:effectivenessDRMJS}
\end{figure}

We have also created a synthetic dataset, DRM (Section ~\ref{sec:datasets}), composed of 1000 Dirichlet distributions with the same dimensions than topics in AIES: $k=44$. Unlike AIES, topic distributions have been randomly generated which imply that the similarity values are not so high: $min=0.06$, $mean=0.18$ and $max=0.61$. Following the same criteria than before (Section ~\ref{sec:threshold}), the similarity threshold is now fixed to 0.34 (Figure ~\ref{fig:similaritiesDRM}). Results \textbf{in terms of \textit{effectiveness}} (Figure ~\ref{fig:effectivenessDRMJS}) show a poor performance of the RDC and CRDC algorithms. The reason is that both are based on the fact that the highest weighted topics are shared between similar distributions. However, this condition is not satisfied when the similarity value between them is low.

\begin{figure}
  \includegraphics[scale=0.27]{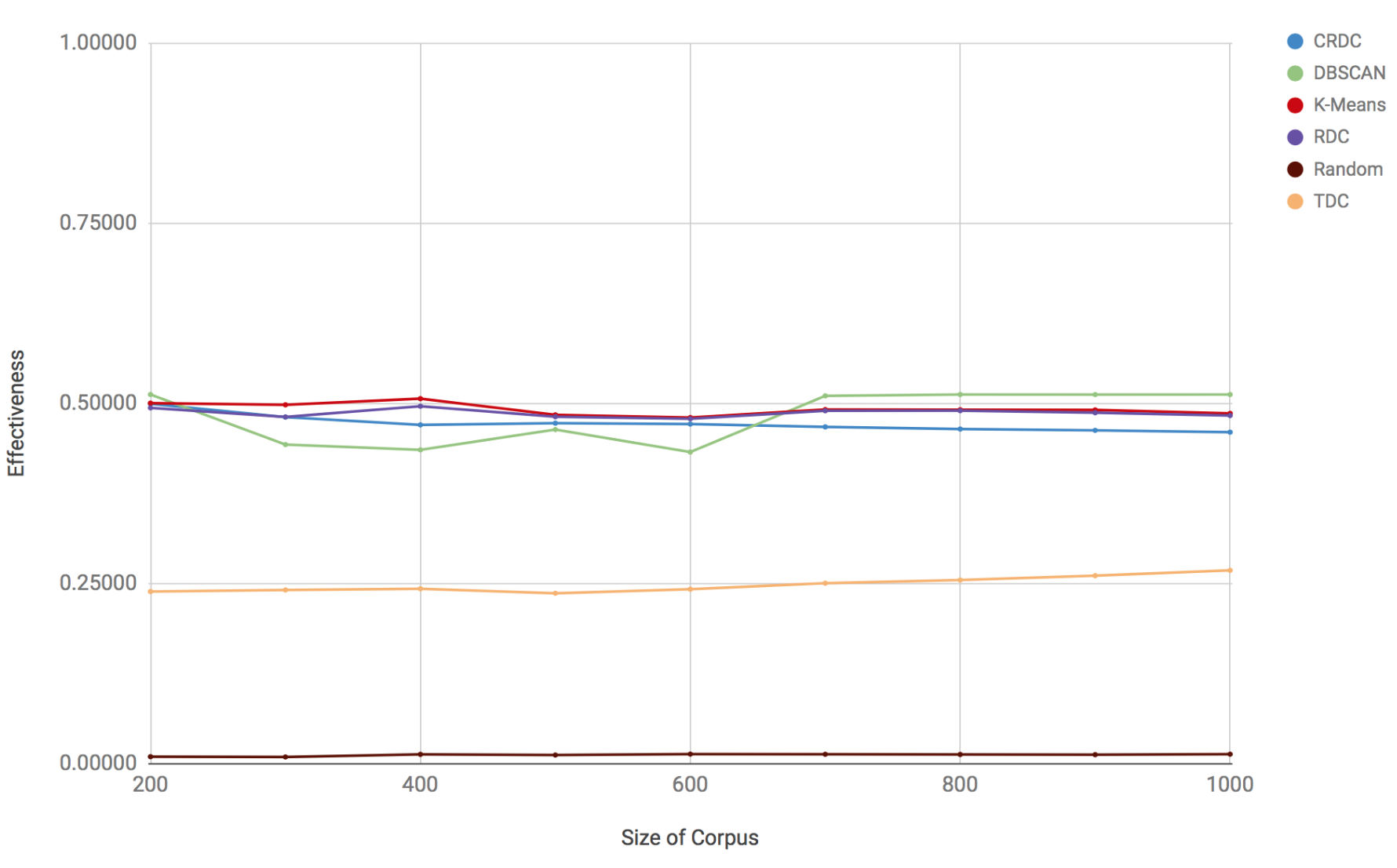}
  \caption{Effectiveness (JS based) in DRM2}
  \label{fig:effectivenessDRM2JS}
\end{figure}

To confirm this behavior, we created a third dataset (DRM2) with the same size but with only 4 dimensions (4 topics). The goal is to achieve more similar distributions than in DRM even though they are also randomly generated. Since the similarity values range from $min=0.04$, $mean=0.34$ to $max=0.99$, the similarity threshold is now fixed to 0.66 (more details in section ~\ref{sec:threshold}). The results (Figure ~\ref{fig:effectivenessDRM2JS}) show an improvement in the accuracy of both the RDC and CRDC algorithms. Although scores are still not as high as for the AIES dataset, the increase compared to the DRM dataset shows that their \textit{precision} and \textit{recall} improve when the similarity threshold is higher. On the other hand, both the DBSCAN and TDC algorithms show similar behavior in both datasets, which means that their performance is not affected by the similarity threshold.

\section{Conclusions and Future Work}
\label{sec:conclusion}
 Processing a continuously growing collection of human generated documents requires techniques that divide the space into smaller regions containing potentially similar documents. Some algorithms in the literature tackle this problem from an unsupervised point of view, but they incur in high temporal costs and may not be suited for the domain being studied.

Three novel unsupervised clustering algorithms, \textit{TDC}, \textit{RDC} and \textit{CRDC}, are described in this paper relying on the distributions inferred from a topic modeling algorithm (LDA). They are presented as a means to identify a smaller set of documents where only the similarity function has to be computed. They leverage on the particular behavior of Dirichlet distributions describing topic distributions, where the highest weighted topics have a high influence on the rest of topics. This also means that given a topic distribution, the relations between their topic weights such as order or trends between them, are more important than the density values.

Although we initially thought that using only a fixed number of topics with higher weights of a topic distribution (\textit{RDC}), or taking into account only the trend changes between the weights of consecutive topics (\textit{TDC}), could be enough to classify similar topic distributions, the results obtained have shown that these properties are not sufficient. Results in terms of \textit{efficiency}, \textit{effectiveness} and \textit{cost} have been shown comparing the proposed algorithms with existing centroid-based and density-based clustering techniques. They reveal that obtaining the most representative topics of a topic distribution  by comparing the sum of their weights with respect to the rest (\textit{CRDC}) is a promising approach, which improves the \textit{efficiency} obtained by other centroid-based and density-based approaches. While \textit{K-Means} takes $O(n^k * \log{n})$ and \textit{DBSCAN} takes $O(n * \log{n})$ time to classify $n$ documents in a collection, the proposed algorithms only take linear time ($O(n)$) because they do not require any other data except their own topic distribution to assign it to a cluster.

A hierarchical approach for RDC algorithm was also considered but it did not produce good results. Hybrid methods combining some of these novel approaches with existing techniques will be performed in future work on the same line.


\bibliographystyle{ACM-Reference-Format}
\bibliography{references}

\end{document}